\documentclass[sigconf,9pt,noacm]{acmart}

\usepackage{afterpage}
\usepackage{algorithmicx}
\usepackage{amsfonts}
\usepackage{amsmath}
\usepackage{amsthm}
\usepackage{array}
\usepackage{blindtext}
\usepackage[linesnumbered,ruled,noend]{algorithm2e}
\usepackage{booktabs}
\usepackage{boxedminipage}
\usepackage{color}
\usepackage{colortbl}
\usepackage{enumitem}
\usepackage{float}
\usepackage{graphics}
\usepackage{graphicx}
\usepackage{indentfirst}
\usepackage{lipsum}
\usepackage{multirow}
\usepackage{picinpar}
\usepackage{placeins}
\usepackage{subfigure}
\usepackage{textcomp}
\usepackage{tikz}
\usepackage[normalem]{ulem}
\usepackage{url}
\usepackage{wrapfig}
\usepackage{balance}
\usepackage{listings}
\usepackage{bookmark}
\usepackage{makecell}

\newtheoremstyle{compact}%
  {0pt}{0pt}
  {\itshape}
  {}
  {\bfseries}
  {.}
  { }
  {}
\theoremstyle{compact}

\allowdisplaybreaks

\SetKwInOut{Input}{Input}
\SetKwInOut{Output}{Output}

\SetKwProg{Fn}{Function}{\string:}{end}

\SetCommentSty{myCommentStyle}

\renewcommand\footnotetextcopyrightpermission[1]{} 
\copyrightyear{}
\acmYear{}
\setcopyright{acmcopyright}\acmConference[]{}{}
\acmBooktitle{}
\acmPrice{}
\acmDOI{}
\acmISBN{}

\newcommand{\tit}[1]{{\textit{#1}}}
\newcommand{\tbf}[1]{{\textbf{#1}}}
\newcommand{\ie}{\textit{i.e.}}
\newcommand{\eg}{\textit{e.g.}}

\newcommand{\etal}{\textit{et al.}}
\newcommand{\METHOD}{TRAM}

\begin{document}

\title{TRAM: Training Approximate Multiplier Structures for Low-Power AI Accelerators}
\author{Chang Meng}
\affiliation{%
  \institution{Eindhoven University of Technology}
  \city{Eindhoven}
  \country{The Netherlands}
}
\email{c.meng@tue.nl}
\author{Hanyu Wang}
\affiliation{%
  \institution{University of California, Los Angeles}
  \city{Los Angeles}
  \country{USA}
}
\email{hanyuwang@g.ucla.edu}
\author{Yuyang Ye}
\affiliation{%
  \institution{Chinese University of Hong Kong}
  \city{Hong Kong}
  \country{China}
}
\email{yuyangye@cuhk.edu.hk}
\author{Mingfei Yu}
\affiliation{%
  \institution{École Polytechnique Fédérale de Lausanne, Lausanne, Switzerland}
  \city{}
  \country{}
}
\email{mingfei.yu@epfl.ch}
\author{Wayne Burleson}
\affiliation{%
  \institution{University of Massachusetts Amherst}
  \city{Amherst}
  \country{USA}
}
\email{burleson@umass.edu}
\author{Giovanni De Micheli}
\affiliation{%
  \institution{École Polytechnique Fédérale de Lausanne, Lausanne, Switzerland}
  \city{}
  \country{}
}
\email{giovanni.demicheli@epfl.ch}


\begin{abstract}

Reducing power consumption in AI accelerators is increasingly important.
Approximate computing can reduce power consumption while keeping the accuracy loss small.
Since multipliers are power-hungry components in AI models,
this paper focuses on synthesizing low-power \tit{approximate multipliers (AxMs)}.
Unlike prior works that design AxMs separately from AI model training, 
we present \tit{\METHOD{}},
which jointly optimizes the AxM structure and AI model parameters to lower power with small accuracy loss.
Experiments show that compared to state-of-the-art AxMs,
\METHOD{} achieves up to 25.05\% AxM power reduction on CNNs with CIFAR-10,
and reduces power by up to 27.09\% on vision transformers with ImageNet.

\end{abstract}



\maketitle

\section{Introduction}\label{sect:intr}

The wide deployment of AI accelerators raises concerns about power consumption
and creates an urgent need for low-power computing solutions~\cite{schwartz2020green}.
Approximate computing
reduces power consumption by allowing inaccuracies in computations,
making it a promising approach to addressing these concerns~\cite{leon2025approximate}.
Since multipliers are among the most power-consuming components in AI accelerators~\cite{armeniakos2022hardware},
this paper studies the automatic synthesis of low-power \tit{approximate multipliers} (\tit{AxM}s).

Many studies have investigated both automatic synthesis and manual design of AxMs~\cite{wu2024survey}.
For example,
Mrazek~\etal{}~\cite{mrazek2017evoapprox8b} proposed a genetic programming-based method to synthesize AxMs
and later extended it to \tit{convolutional neural networks (CNNs)}~\cite{mrazek2020libraries}.
Xiao~\etal{}~\cite{xiao2022opact} formulated AxM synthesis as an integer programming problem and produced low-cost AxMs.
Hu~\etal{}~\cite{hu2024configurable} manually designed AxMs for CNNs using partial product speculation.
Furthermore, approximate logic synthesis tools, such as those proposed in~\cite{wang2023dasals,ma2021approximate,meng2026simulation}, can synthesize AxMs as well.

However, all the above methods overlook the specific context of AI models, which can lead to suboptimal results when deploying AxMs in accelerators.
First, many existing methods do not consider the data distribution used by AI models.
For instance, Xiao~\etal{}~\cite{xiao2022opact} assumed a uniform input distribution,
while real data distributions vary across layers.
Ignoring this variability can lead to suboptimal designs.
Second, most existing works design or synthesize AxMs using local error metrics such as error rate or error distance.
However, a small local error does not always translate into a small final accuracy loss in the AI model.





\begin{figure}[t]
    \centering
    \includegraphics[width=1.00\columnwidth]{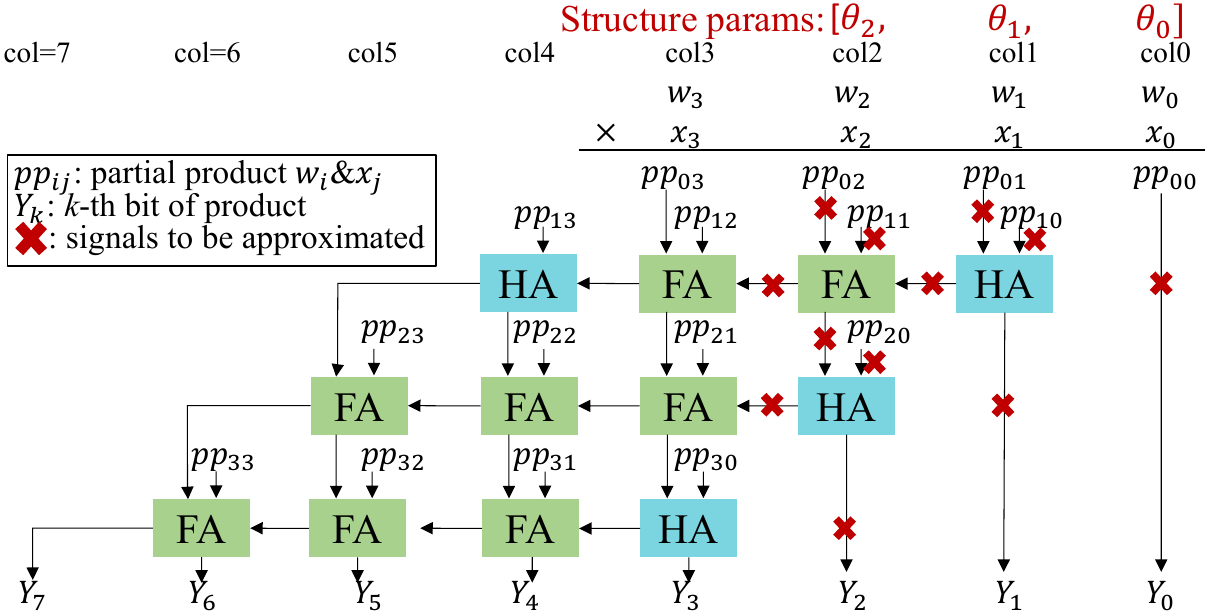}
    \caption{A 4-bit unsigned array multiplier.
    The red crosses indicate candidate signals that may be approximated.
    The structure parameter $\theta_i$ controls the approximation degree of the $i$-th accumulation column.
    We assume that at most $P=3$ columns can be approximated.
    }
    \label{fig:appmult}
    \Description{}
\end{figure}

\begin{figure*}[t]
    \centering
    \includegraphics[width=0.65\textwidth]{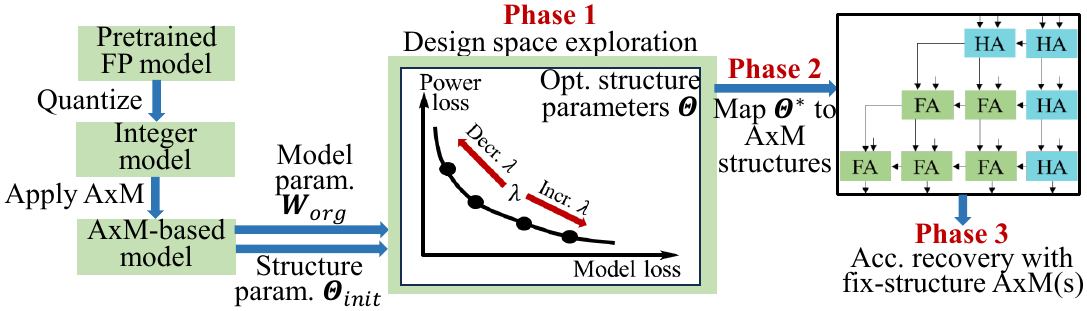}
    \caption{\METHOD{} framework overview.}
    \label{fig:overall}
    \Description{}
\end{figure*}

To address these issues,
we present \tit{\METHOD{}},
a hardware-software co-optimization framework that \underline{TR}ains \underline{A}pproximate \underline{M}ultiplier structures for low-power AI accelerators.
\METHOD{} formulates AxM synthesis as a joint optimization problem that updates the AxM structure and AI model parameters together during training.
By using real training data, \METHOD{} captures the statistics seen by each multiplier and optimizes both the multiplier structure and model parameters with respect to the final accuracy loss.
Our contributions are summarized as follows:
\begin{itemize}
    \item 
    We introduce a parameterization of the AxM structure,
    in which each column in the compressor tree is assigned a continuous \tit{structure parameter} that controls the approximation degree of the column.
    These parameters are optimized using gradient descent.

    \item We devise an analytic power model that estimates multiplier power from the structure parameters 
    and provides useful hardware-aware guidance during training.

    \item We propose an efficient mapping method that converts the optimized structure parameters into concrete AxM designs.

\end{itemize}%

Experimental results show that, compared to state-of-the-art AxM designs,
\METHOD{} reduces AxM power by up to 25.05\% on CNNs with CIFAR-10 at the same accuracy level, 
and by 27.09\% on vision transformers with ImageNet.
Since \METHOD{} allows different structure parameters for different model layers, 
it naturally supports layer-wise application of different AxMs. 
Compared to the state-of-the-art layer-wise AxM exploration methods, 
\METHOD{} reduces AxM energy by 40.86\%.
Our work is open source and available at \url{https://github.com/changmg/TRAM}.

The rest of this paper is organized as follows.
Section~\ref{sect:prel} describes preliminaries.
Sections~\ref{sect:method}--\ref{sect:extract} detail the \METHOD{} framework.
Section~\ref{sect:result} discusses the experimental results.
Section~\ref{sect:concl} concludes this paper.

\section{Integer Multiplier Preliminaries}\label{sect:prel}

This paper focuses on \emph{unsigned integer multipliers}
that are widely used in AI accelerators~\cite{simon2021exact,jain2022learning,meng2025gradient,zheng2022heam}.
Hereafter, we refer to unsigned integer multipliers simply as multipliers.
AxMs are usually obtained by modifying \tit{accurate multipliers (AccMuls)}.
A $B$-bit AccMul computes the exact product of two unsigned integer inputs $W$ and $X$,
which are represented in binary as $W = w_{B-1}w_{B-2}\ldots w_0$ 
and $X = x_{B-1}x_{B-2}\ldots x_0$.

The multiplier contains $2B$ accumulation columns.
The $c$-th column accumulates the partial products as $S_c = \sum_{i=0}^{c} pp_{i, c-i}$,
where $pp_{i,j} = w_i \cdot x_j$ is the partial product of $w_i$ and $x_j$,
and $0 \le c \le 2B - 1$ is the column index.
The final product is obtained by summing the weighted accumulation results from all columns as $Y = \sum_{c=0}^{2B-1} S_c \cdot 2^c$.
For example,
Fig.~\ref{fig:appmult} shows a 4-bit array multiplier with 8 accumulation columns.
Each column generates partial products and accumulates them using \tit{half adders (HAs)} and \tit{full adders (FAs)}.
Approximation can be introduced to the partial products or to the sum and carry-out signals of the half adders and full adders in these columns, as shown by the red crosses in Fig.~\ref{fig:appmult}.

To evaluate the accuracy of a $B$-bit AxM,
common error metrics include \tit{error rate (ER)}, \tit{normalized mean error distance (NMED)}, and \tit{maximum error distance (MaxED)}~\cite{jiang2020approximate}, defined as
\[
\tit{ER} = \sum_{1\leq i \leq 2^{2B}:Y^{(i)} \neq Y_\tit{acc}^{(i)}}{p_i}, \quad
\tit{NMED} = \sum_{i=1}^{2^{2B}}{\frac{\left|Y^{(i)} - Y_\tit{acc}^{(i)}\right|\cdot p_i}{2^{2B} - 1}}, \quad
\]
\[
\tit{MaxED} = \max_{1\leq i \leq 2^{2B}}{\left|Y^{(i)} - Y_\tit{acc}^{(i)}\right|}.
\]
where $Y^{(i)}$ and $Y_\tit{acc}^{(i)}$ are the outputs of the AxM and the AccMul under the $i$-th input combination,
$p_i$ is the probability of the $i$-th input combination,
and $2^{2B}$ is the total number of input combinations.

\section{\METHOD{} Overview and Multiplier Structure Parameterization}\label{sect:method}


\begin{figure*}[t]
    \centering
    \includegraphics[width=0.7\textwidth]{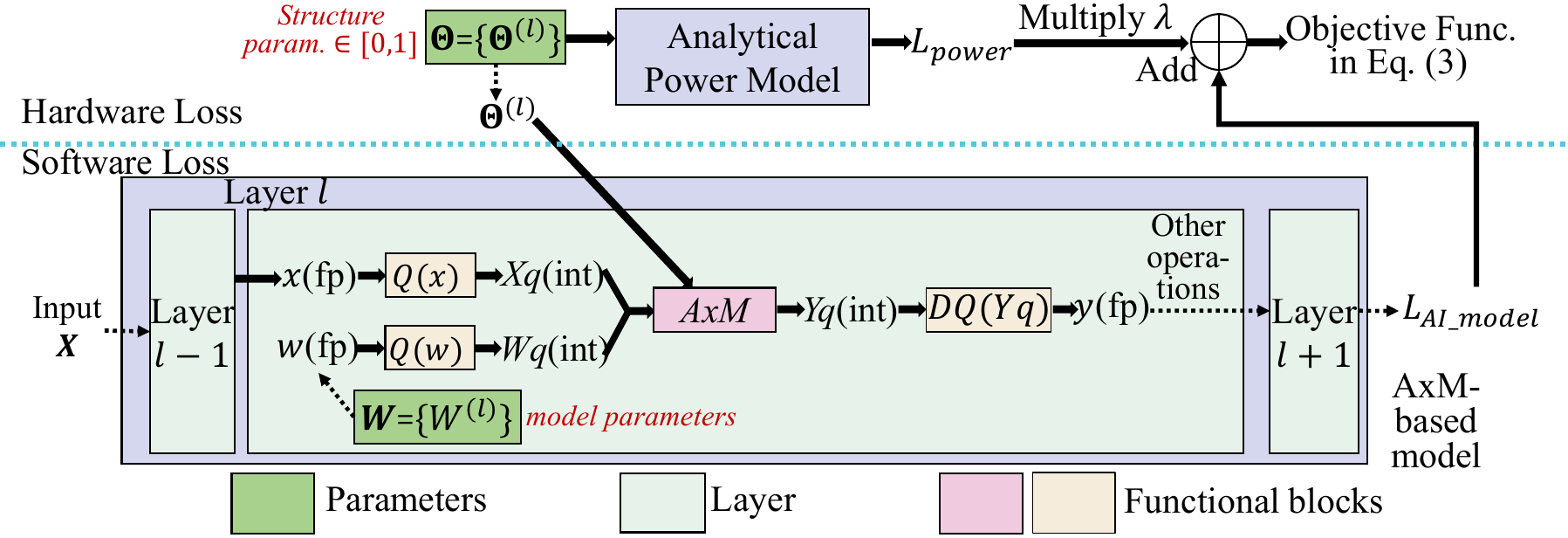}
    \caption{Dataflow for computing the objective function in Eq.~\eqref{eq:objective}.
    The upper part computes the power loss $\mathcal{L}_\tit{power}$ (Section~\ref{subsect:power_loss}),
    and the lower part computes the AI model loss  $\mathcal{L}_\tit{AI\_model}$ (Section~\ref{subsect:ai_model_loss}).
    }\label{fig:data-flow}
    \Description{}
\end{figure*}

\subsection{\METHOD{} Framework Overview}\label{subsect:overview}

\METHOD{} aims to generate low-power AxMs for AI accelerators.
The overall flow of \METHOD{} is shown in Fig.~\ref{fig:overall}.
It starts from a pretrained floating-point AI model,
which is then quantized into an integer model.
To further reduce power, 
AxMs replace the AccMuls in the quantized model.
To explore the AxM design space,
we represent AxM structures using the \tit{structure parameters} collected in $\Theta$.
Changing the AxM structure corresponds to updating $\Theta$.
The detailed parameterization of $\Theta$ is presented in Section~\ref{subsect:indicator}.
Based on this parameterization, we propose a three-phase method to generate low-power AxMs for high-accuracy AI models:

\noindent\tbf{Phase 1.} Design space exploration (details in Section~\ref{sect:loss_computation}).
This phase explores the AxM design space defined by $\Theta$ through model retraining and balances power and accuracy.

\noindent\tbf{Phase 2.}
AxM structure mapping (details in Section~\ref{sect:extract}).
This phase maps the optimized continuous structure parameters in $\Theta^{*}$ from phase 1 to specific AxM structures for each layer of the model.

\noindent\tbf{Phase 3.}
Accuracy recovery.
After mapping the structure parameters to AxM structures in phase 2,
we apply these AxMs to the AI model and retrain it to recover the accuracy.



\subsection{AxM Structure Parameterization}\label{subsect:indicator}

We parameterize the AxM structure using continuous structure parameters,
enabling gradient-based AxM structure optimization through model retraining.
Let $\Theta{=}\{\Theta^{(l)}\} (1{\le}l{\le}L)$ denote the collection of structure parameters for all $L$ layers in the model.
We assume each layer uses one AxM structure for all multiplications in that layer,
described by $\Theta^{(l)}$.
$\Theta^{(l)}$ has $P$ parameters:
$\Theta^{(l)}{=}[\theta^{(l)}_0, \theta^{(l)}_1, \ldots, \theta^{(l)}_{P-1}]$ (see Fig.~\ref{fig:appmult}),
where $\theta^{(l)}_c$ ($0{\le}c{\le}P{-}1$) describes the approximation degree of column $c$.
Here, $P$ is a user-defined maximum number of columns that can be approximated.

Each $\theta^{(l)}_c \in [0, 1]$ is a continuous structure parameter that controls the approximation degree of column $c$ in the AxM of layer $l$.
A value of 0 means the column is kept fully accurate, 
while a value of 1 means that the column is entirely removed.
An intermediate value $0{<}\theta^{(l)}_c{<}1$ represents a partial approximation, 
where only a subset of the partial products or compressors in column $c$ is removed.

Next, we explain how the structure parameters control the functional behavior of the AxM.
For the $l$-th layer, 
the approximation error of the $c$-th accumulation column is defined as
    $E_c = \theta^{(l)}_c \cdot S_c = \theta^{(l)}_c \cdot \sum_{i=0}^{c} pp_{i,c-i}$,
where $S_c$ is the exact accumulation result of column $c$.
The total approximation error over all $P$ approximated columns is
computed by summing the column errors multiplied by their weights $2^c$, \ie{},
$E_{total} = \sum_{c=0}^{P-1} E_c \cdot 2^c$.
The AxM output is then obtained by subtracting this error from the exact product:
\begin{equation}\label{eq:approx_output}
Y = WX - E_{total} = WX - \sum_{c=0}^{P-1} \theta^{(l)}_c \cdot S_c \cdot 2^c,
\end{equation}%
where $WX$ is the exact product of $W$ and $X$.
Eq.~\eqref{eq:approx_output} enables
smooth adjustment of the approximation degree in each column $c$ by varying the structure parameter $\theta^{(l)}_c$ in the range $[0, 1]$.
A larger $\theta^{(l)}_c$ leads to a larger approximation error
and reduces power consumption,
while a smaller $\theta^{(l)}_c$ reduces the error and increases power.
This formulation can be extended beyond array multipliers. 

\section{Phase 1: Design Space Exploration through AI Model Retraining}\label{sect:loss_computation}

\subsection{Problem Formulation}\label{subsect:problem_formulation}
The structure parameters $\Theta$ define the AxM design space.
Different choices of $\Theta$ lead to AxMs with different power consumption and different AI model accuracy.
To balance power consumption and accuracy,
we formulate the following optimization problem:
\begin{equation}\label{eq:objective}
    \min_{\Theta, \mathbf W} 
    \left(\mathcal{L}_\tit{power}(\Theta) \cdot \lambda
    + 
    \mathcal{L}_\tit{AI\_model}(\Theta, \mathbf W, \mathbf X)\right).
\end{equation}%
Eq.~\eqref{eq:objective} consists of two loss terms:
the power loss $\mathcal{L}_\tit{power}$ 
and
the AI model loss $\mathcal{L}_\tit{AI\_model}$.
$\mathcal{L}_\tit{power}$ maps the structure parameters $\Theta$ to the total power consumed by all AxMs in the AI accelerator.
$\mathcal{L}_\tit{AI\_model}$ is the original model loss 
(\eg{}, \tit{cross-entropy loss} for classification)
and depends on the structure parameters $\Theta$, 
the model weights $\mathbf W$,
and the inputs $\mathbf X$.
A trade-off parameter $\lambda$ is introduced to balance these two losses.
Increasing $\lambda$ gives more weight to $\mathcal{L}_\tit{power}$,
which lowers power consumption but increases model loss (\ie{}, lower model accuracy).
By tuning $\lambda$,
we can explore different power-accuracy trade-offs.

We solve the optimization problem in Eq.~\eqref{eq:objective} through model retraining.
During retraining,
the dataflow for computing the objective in Eq.~\eqref{eq:objective} is shown in Fig.~\ref{fig:data-flow}.
The upper part computes $\mathcal{L}_\tit{power}$
and the lower part computes $\mathcal{L}_\tit{AI\_model}$.
Sections~\ref{subsect:power_loss} and~\ref{subsect:ai_model_loss} describe how $\mathcal{L}_\tit{power}$
and $\mathcal{L}_\tit{AI\_model}$ are computed within this dataflow.

\subsection{Computation of Power Loss}\label{subsect:power_loss}

The power loss $\mathcal{L}_\tit{power}(\Theta)$ estimates the power consumed by all AxMs in all layers of the AI model as follows:
\begin{equation}\label{eq:power_loss}
    \begin{aligned}
    & \mathcal{L}_\tit{power}(\Theta) = \sum_{l=1}^{L} f_\tit{power}(\Theta^{(l)})\times \frac{\text{\#mults at layer }l}{\text{\#mults in all layers}},\\
    \end{aligned}
\end{equation}%
where $\Theta = \{\Theta^{(l)}\} (1\leq l\leq L)$ is the collection of structure parameters for the $L$ layers.
Here, $\Theta^{(l)}$ is the structure parameter vector for layer $l$,
and $f_\tit{power}(\Theta^{(l)})$ denotes the power of the AxM used in that layer.
Eq.~\eqref{eq:power_loss} is a weighted sum of the AxM power over all layers.
The weight for layer $l$ is the ratio between its number of multiplication operations and the total multiplication count of the whole model.
This weight is computed from the multiplication counts of the model layers.
This weight approximates the fraction of total inference latency spent on layer $l$.
Thus, the weighted sum estimates the time-averaged AxM power during inference.

We propose an analytical method to compute $f_\tit{power}(\Theta^{(l)})$.
We first estimate the power of an AccMul as follows:
\begin{equation}\label{eq:power_model}
    \tit{Power}_\tit{AccMul} = \sum_{c=0}^{2B-1} \tit{Power}_c
    = \sum_{c=0}^{2B-1} \sum_{k=1}^{K_c} \tit{cost}_{c,k} \cdot N_{c,k}.
\end{equation}
Here, $B$ is the multiplier bit-width,
which results in $2B$ accumulation columns in the AccMul.
$\tit{Power}_c$ is the power of the $c$-th accumulation column,
computed by summing the power of all $K_c$ component types in that column.
Example component types include logic AND gates for partial-product generation and various compressors for accumulation.
$\tit{cost}_{c,k}$ is the power of a type-$k$ component in column $c$,
pre-characterized using the standard cell library.  
$N_{c,k}$ is the number of type-$k$ components in column $c$ of the AccMul.
For example, 
consider the array multiplier in Fig.~\ref{fig:appmult}.
Column 2 contains 3 AND gates for partial-product generation, 1 HA and 1 FA for accumulation.
Thus, column 2 has $K_2=3$ component types,
labelled as type 1 (AND gate), type 2 (HA), and type 3 (FA).
The number of each component type is $N_{2,1}=3$, $N_{2,2}=1$, and $N_{2,3}=1$.
Assume the power of each type is $\tit{cost}_{2,1}=1$, $\tit{cost}_{2,2}=2$, and $\tit{cost}_{2,3}=3$.
Then, the power consumption of column 2 is $\tit{Power}_2 = 1\times 3 + 2\times 1 + 3\times 1 = 8$.

Recall that $\Theta^{(l)}$ contains $P$ structure parameters $\theta^{(l)}_c (0 \le c \le P-1)$ for layer $l$.
From Eq.~\eqref{eq:approx_output},
$\theta^{(l)}_c\in [0,1]$ specifies the fraction of logic components removed from the $c$-th column of layer $l$.
Based on this, we estimate the normalized AxM power for layer $l$ as follows:

\begin{equation}\label{eq:power_model_am}
    f_\tit{power}(\Theta^{(l)}) = \frac{\tit{Power}_\tit{AccMul} - \sum_{c=0}^{P-1}  \theta^{(l)}_c \cdot \tit{Power}_c}{\tit{Power}_\tit{AccMul}},
\end{equation}%
where $P$ is the maximum number of approximated columns,
and $\tit{Power}_c$ is the power of the $c$-th accurate accumulation column.
When $\theta^{(l)}_c$ is close to 0,
most components in column $c$ are preserved and the power remains high.
Scaling $\tit{Power}_c$ by $\theta^{(l)}_c$ provides an estimate of the power reduction gained from approximating that column.
When $\theta^{(l)}_c$ is close to 1,
most components in column $c$ are removed
and the power becomes low.

\subsection{Computation of the AI Model Loss}\label{subsect:ai_model_loss}

The AI model loss function $\mathcal L_\tit{AI\_model}$,
such as cross-entropy loss for classification tasks,
measures the difference between the model output and the ground truth.
It depends on the structure parameters $\Theta$, the model parameters $\mathbf W$,
and the input data $\mathbf X$.
For illustration, we present the computation of $\mathcal{L}_\tit{AI\_model}$ in the context of a CNN.


As shown in the bottom part of Fig.~\ref{fig:data-flow},
the forward propagation of $\mathcal{L}_\tit{AI\_model}$ 
processes the layers of the model in order.
For CNNs, we replace the accurate multiplications in convolutional layers with AxMs,
while for transformer-based models,
we replace all linear layers
in attention and feed-forward blocks with AxMs.
In what follows,
we first present how to simulate the AxM behavior.
Since the AxM requires quantized inputs, we then present how to simulate the quantization process.

\tbf{AxM Simulation}:
As shown in the center of Fig.~\ref{fig:data-flow},
at the $l$-th layer,
an AxM takes the integer activation $X_q$, 
the integer weight $W_q$,
and the structure parameter vector $\Theta^{(l)}$ as inputs.
The AxM produces a quantized integer output $Y_q$.
As described in Section~\ref{subsect:indicator},
$\Theta^{(l)}$ is a vector of real values in $[0,1]$ with length $P$,
where $P$ is the maximum number of approximated columns.
Using Eq.~\eqref{eq:approx_output}, 
we can express the AxM computation in closed form as:
\begin{equation}\label{eq:reconf-axmul}
    \begin{aligned}
    Y_q = W_q X_q - \sum_{c=0}^{P-1}\left[ 2^{c}\cdot \theta_{c}^{(l)} \cdot \sum_{i=0}^{c} \left( W_q[i] \cdot X_q[c - i] \right)\right], \\
    \end{aligned}
\end{equation}%
where $W_q[i]$ is the $i$-th bit of $W_q$, 
and $X_q[j]$ is the $j$-th bit of $X_q$.
The term ($W_q[i] \cdot X_q[c-i]$) is the partial product of $W_q[i]$ and $X_q[c-i]$.
The sum of partial products in column $c$ is scaled by the continuous structure parameter $\theta_{c}^{(l)} \in [0, 1]$.
If $\theta_{c}^{(l)}=0$,
the $c$-th accumulation column is kept exactly.
If $\theta_{c}^{(l)}=1$,
the $c$-th accumulation column is fully removed.
If $0 < \theta_{c}^{(l)} < 1$,
logic components in column $c$
are partially removed,
and the corresponding accumulation error is estimated by scaling the exact accumulation result using $\theta_{c}^{(l)}$.

\tbf{Quantization Simulation}: 
Since the inputs and outputs of an AxM are integers,
quantization is required before the AxM operation.
We apply the traditional \tit{fake quantization} technique~\cite{jacob2018quantization} during training.
The quantization functions $Q(x)$ and $Q(w)$, and the dequantization function  $DQ(Y_q)$ in Fig.~\ref{fig:data-flow} follow~\cite{shao2023omniquant}.
\section{Phase 2: AxM Structure Mapping}\label{sect:extract}





Phase~1 produces a set of continuous structure parameters $\Theta^{*}=\{\Theta^{*(1)}, \Theta^{*(2)},\ldots,\Theta^{*(L)}\}$,
where $\Theta^{*(l)}$ 
corresponds to the AxM used in layer $l$.
Let $\Theta^{*(l)}=[\theta^{*}_0,\theta^{*}_1,\ldots,\theta^{*}_{P-1} ]$ (layer index $l$ omitted for brevity),
and $P$ is the maximum number of columns that can be approximated.
These continuous structure parameters $\theta^{*}_c$
cannot be directly implemented in hardware.

The goal of phase 2 is to map $\Theta^{*(l)}$ into a concrete AxM netlist used in each layer $l$
so that the resulting circuit behaves as closely as possible to the behavior implied by $\Theta^{*(l)}$.
Specifically, during training, $\Theta^{*(l)}$
controls the amount of error added to each column of the AxM.
The hardware mapping aims to reproduce this same error behavior.
To guide this mapping, 
we compute the expected AxM output under $\Theta^{*(l)}$ using the closed-form model in Eq.~\eqref{eq:reconf-axmul}.
For all input combinations,
we compute the reference output:
\begin{equation}
\label{eq:yref}
Y_{\mathrm{ref}}
= W_q X_q
- \sum_{c=0}^{P-1}
  \Bigg[
    2^{c}\,\theta^{*}_{c}
    \sum_{i=0}^{c}
      \big(W_q[i]\cdot X_q[c-i]\big)
  \Bigg].
\end{equation}%
Eq.~\eqref{eq:yref} defines the target behavior that the hardware should match.
Thus, given $\Theta^{*(l)}$, our task is to construct an AxM whose output function approximates $Y_{\mathrm{ref}}$ as closely as possible.

To obtain such an AxM,
we begin with an AccMul.
In our implementation, the initial multiplier is an array-based AccMul.
For each accumulation column $c$, the candidates for approximation are
the sum and carry outputs of compressors such as HAs and FAs in column $c$.
Each candidate can be tentatively replaced by constant~$0$.
Such replacements act as discrete forms of the continuous error implied by $\theta^{*(l)}_c$.

Inspired by existing approximate logic synthesis methods
that assess local replacements using the errors they induce~\cite{meng2020alsrac,meng2023hedals,hashemi2018blasys,meng2024efficient},
we define an error-based metric to evaluate whether a constant replacement of a candidate signal is beneficial.
Let $Y^{\text{curr}}_{\mathrm{circ}}$ be the output of the current circuit and $Y^{\text{new}}_{\mathrm{circ}}$ the output after a tentative replacement.
We evaluate the errors of $Y^{\text{curr}}_{\mathrm{circ}}$ and $Y^{\text{new}}_{\mathrm{circ}}$ with respect to $Y_{\mathrm{ref}}$ using the same input patterns.
A replacement is accepted only if it strictly decreases the \tit{mean squared error (MSE)} from the reference:
$\operatorname{MSE}\!\big(Y^{\text{new}}_{\mathrm{circ}},\,Y_{\mathrm{ref}}\big)
<
\operatorname{MSE}\!\big(Y^{\text{curr}}_{\mathrm{circ}},\,Y_{\mathrm{ref}}\big).$


Based on this evaluation metric,
we propose a greedy column-wise mapping flow.
We traverse the multiplier from the least significant column to the most significant column.
For each column, all approximation candidates are considered before moving to the next column.
Each candidate is tentatively replaced by constant~$0$ and evaluated using the $\operatorname{MSE}$ criterion.
If a tentative substitution reduces the MSE, it is permanently applied to the circuit. Otherwise, the tentative substitution is undone.
After processing all columns,
we obtain a circuit whose output is close to the reference output $Y_\tit{ref}$.
Finally, the resulting AxM netlist is emitted as Verilog.
Note that this procedure always yields a feasible design, since it starts from an AccMul and only applies constant replacements.


\vspace{-1em}
\section{Experimental Results}\label{sect:result}

We implement the \METHOD{} framework using PyTorch 2.4~\cite{paszke2019pytorch} and test it on a single NVIDIA A100 GPU.
Experiments are conducted with CNN and ViT models on the CIFAR-10~\cite{krizhevsky2009learning} and ImageNet~\cite{deng2009imagenet} datasets.
We evaluate two quantization schemes: \tit{w8a8} (8-bit weights and activations) and \tit{w4a4} (4-bit weights and activations),
covering both standard and low-bitwidth regimes commonly used in AI accelerators.
The initial w8a8-quantized model is obtained through post-training quantization,
while the initial w4a4-quantized model is prepared using quantization-aware training.
Channel-wise quantization is applied to the weights,
and layer-wise quantization is applied to the activations.

Unless otherwise specified,
all experiments use the following default settings.
We use a batch size of 256 and the SGD optimizer with momentum 0.9 and weight decay 5e-4.
The retraining epochs for design space exploration (phase 1) and accuracy recovery (phase 3) are both set to 10, which we found sufficient for convergence.
Phase 1 uses a fixed learning rate of 5e-4,
while phase 3 uses a cosine annealing schedule that decreases the learning rate from 5e-4 to 0.
The parameter $P$ (see Section~\ref{subsect:indicator}),
the maximum number of accumulation columns allowed to be approximated, is set to 8.
Initially, the AxM in each layer $l$ removes 4 accumulation columns
by setting $\theta^{(l)}_c = 1$ for $c=0,1,2,3$ and $\theta^{(l)}_c = 0$ for all other columns,
serving as a moderate starting point for the optimizer.
To evaluate area, delay, and power,
we synthesize the AxMs using a commercial logic synthesis tool
with the ASAP 7nm standard cell library~\cite{clark2016asap7}.
Power measurements assume a 100 MHz clock frequency. 


We evaluate \METHOD{} under two scenarios:
1) all layers in the AI model share the same AxM type, to compare individual AxM designs against baselines, and
2) different layers can use different AxM types, to evaluate the benefit of layer-wise optimization.

\vspace{-1em}
\subsection{Experiments with a Single AxM Type}\label{subsect:exp-axmul}


This set of experiments assumes that all layers in the AI model use the same type of AxM.
To achieve this in \METHOD{},
we restrict the structure parameters of all layers to be identical,
\ie{},
$\Theta^{(1)}=\Theta^{(2)}=\ldots=\Theta^{(L)}$,
where $L$ is the number of layers.
The baseline AxM designs are ``Evo'', ``OPACT'', and ``AMPPS'',
taken from~\cite{mrazek2020libraries},~\cite{xiao2022opact}, and~\cite{hu2024configurable}, respectively.
The tested 8-bit unsigned multipliers and their errors, areas, and delays are listed in Table~\ref{tab:axmul_info}.
The ER, NMED, and MaxED metrics of the AxMs (see Section~\ref{sect:prel}) are measured by enumerating all possible input combinations under a uniform distribution.
We use the open-source tool in~\cite{meng2024vacsem} to obtain the
ER and NMED metrics.
For a fair comparison,
for each Evo and OPACT AxM,
we use the same number of epochs of AxM-aware retraining as \METHOD{}.
We reimplement this retraining process following the method in~\cite{danopoulos2022adapt}.
However,
the \tit{AMPPS} AxMs include an encoding process
that is not supported by the existing AxM-aware retraining methods.
Thus, we do not perform retraining for the AMPPS AxMs,
and we directly take the final CNN accuracy reported in~\cite{hu2024configurable}.




\begin{table}[tb]
\centering
\tabcolsep=4pt
\caption{Tested 8-bit unsigned multipliers.}\label{tab:axmul_info}
\begin{tabular}{@{}crrrrrr@{}}
\toprule
Multiplier    & \makecell{Area\\/$\mu m^2$}  & \makecell{Delay\\/ps} & \makecell{Power\\/mW}         & ER/\%       & NMED/\%      & MaxED       \\ \midrule
AccMul       & 27.2                       & 496.5    & 0.0031   & 0.0   & 0.000   & 0     \\
OPACT\_1     & 18.6                       & 499.6    & 0.0019   & 23.0  & 0.022   & 516   \\
OPACT\_18 & 14.1                       & 495.1    & 0.0011   & 51.0  & 0.619   & 7684  \\
Evo\_0AB     & 21.3                       & 499.6    & 0.0022   & 97.7  & 0.057   & 115   \\
Evo\_1DMU    & 16.9                       & 499.2    & 0.0015   & 66.0  & 0.650   & 4084  \\
Evo\_GJM     & 12.1                       & 499.5    & 0.0010   & 74.9  & 1.543   & 9124  \\
AMPPS\_S2    & 17.0                       & 499.9    & 0.0019   & 96.3  & 0.076   & 417   \\
AMPPS\_S3    & 18.9                       & 497.9    & 0.0020   & 95.9  & 0.073   & 417   \\
AMPPS\_S4    & 20.8                       & 499.7    & 0.0022   & 95.7  & 0.070   & 417   \\ \bottomrule
\end{tabular}
\end{table}

\subsubsection{Experiments on CIFAR-10}


\begin{figure}[tb]
    \centering
    \includegraphics[width=1.0\columnwidth]{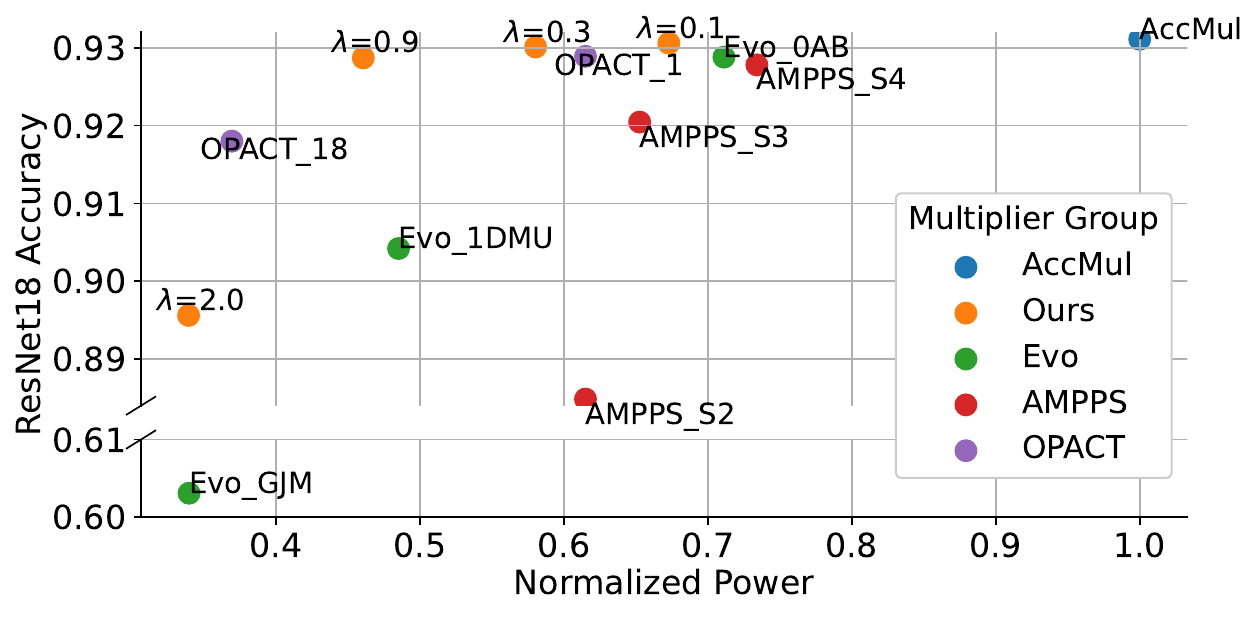}
    \includegraphics[width=1.0\columnwidth]{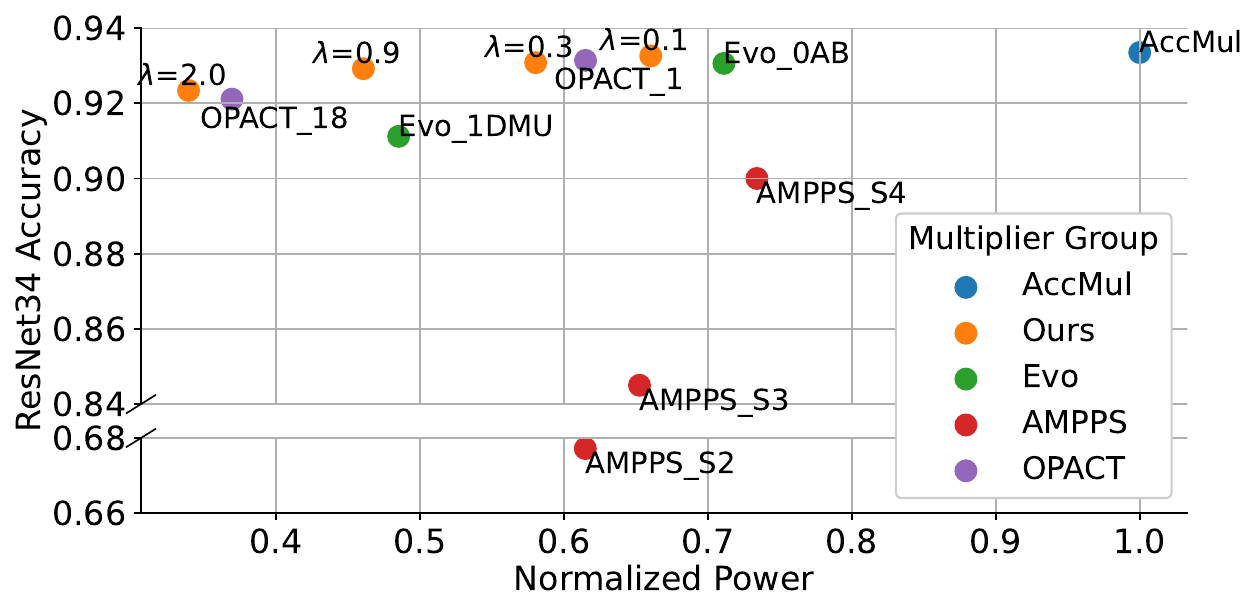}
    \caption{Comparison of final accuracy and AxM power consumption using different 8-bit multipliers on CIFAR-10.
    }\label{fig:w8a8-cifar10}
    \Description{}
\end{figure}

This set of experiments compares the CNN accuracy and the AxM power consumption on CIFAR-10.
We first use the w8a8 quantization scheme and test the ResNet18 and ResNet34 models.
Fig.~\ref{fig:w8a8-cifar10} plots the final model accuracy versus the AxM power consumption for the tested AxMs on ResNet18 and ResNet34.
Each figure includes the 8-bit AccMul result in the upper right corner for reference.
A better AxM achieves higher accuracy and lower power, 
which appears toward the upper left region of the plot.
We vary the trade-off parameter $\lambda$ in Eq.~\eqref{eq:objective} to generate different AxM designs using \METHOD{}.
As $\lambda$ increases,
both the AxM power consumption and the accuracy decrease.
The results show that \METHOD{} achieves a better trade-off between accuracy and power consumption than 
the Evo, AMPPS, and OPACT AxMs for both ResNet18 and ResNet34 models.
For example, compared with OPACT\_1 on ResNet18, \METHOD{} with $\lambda=0.9$ reaches a similar accuracy,
while reducing power by 15.44\%.
On ResNet34,
\METHOD{} with $\lambda=0.9$ reaches a similar accuracy as Evo\_0AB and OPACT\_1,
while reducing power by 25.05\% and 15.43\%, respectively.

As for runtime,
\METHOD{} includes two retraining phases (phases 1 and 3 in Fig.~\ref{fig:overall}) and one AxM structure mapping phase (phase 2).
Phase 2 uses a greedy algorithm
and takes less than 1 minute to generate an 8-bit AxM.
The phase 2 runtime is independent of the CNN model size.
With 20 training epochs in total (10 epochs for phase 1 and 10 epochs for phase 3),
\METHOD{} takes about 17 and 30 minutes for ResNet18 and ResNet34, respectively.

\begin{figure}[tb]
    \centering
    \includegraphics[width=0.82\columnwidth]{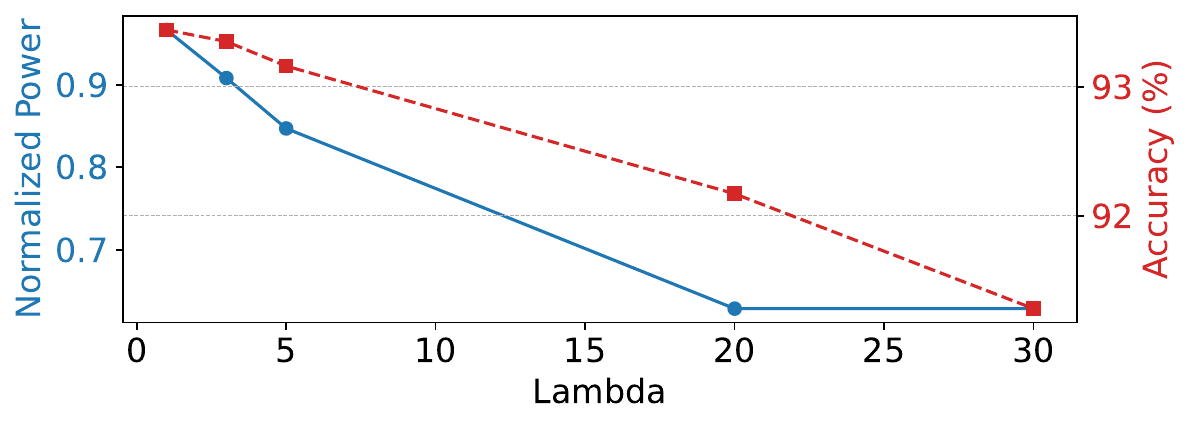}
    \caption{Impact of $\lambda$ on DenseNet161 accuracy and AxM power under w4a4. Power is normalized to the 4-bit AccMul.}
    \label{fig:lambda-w4a4}
    \Description{}
\end{figure}

We also evaluate the impact of $\lambda$ on the DenseNet161 model under the w4a4 quantization scheme.
The maximum number of approximated columns, $P$, is set to 4.
The results are shown in Fig.~\ref{fig:lambda-w4a4}.
Similarly to previous results,
increasing $\lambda$ reduces the AxM power and model accuracy.
The runtime of \METHOD{} for DenseNet161 under w4a4 is about 1.66 hours for 20 training epochs (phases 1 and 3).

\vspace{-1em}
\subsubsection{Experiments on ImageNet}

This set of experiments evaluates \METHOD{} 
on the ImageNet dataset~\cite{deng2009imagenet}.
The baseline method is TransAxx~\cite{danopoulos2025transaxx},
an AxM-aware retraining method for transformer models.
TransAxx tested the power-accuracy trade-off of several 8-bit AxMs from the EvoApprox library~\cite{mrazek2020libraries},
and we compare the \METHOD{}-generated AxMs with these designs.

We refer to the TransAxx training settings,
which use up to 15 epochs for AxM-aware training.
We perform phase 1 (design space exploration) of \METHOD{} for 5 epochs and phase 3 (accuracy recovery) for 10 epochs.
Following the TransAxx setup,
we sample 100,000 images from the 1.28 million training images for efficient training,
and use the full validation set for evaluation.
We use the Adam optimizer with a learning rate of 5e-5 for both phases
and set the batch size to 64.
Note that \METHOD{} uses channel-wise weight quantization and layer-wise activation quantization,
whereas TransAxx uses layer-wise quantization for both weights and activations.
Therefore, the comparison with TransAxx should be viewed as a reference rather than a strictly matched apples-to-apples comparison.
To save runtime, we set the parameter $P$,
which is the maximum number of approximated columns, to 6.

\begin{table}[tb]
\centering
\caption{Results on ImageNet with vision transformers.}
\label{tab:vit}
\begin{tabular}{@{}llrr@{}}
\toprule
Model               & AxM & Accuracy    & \makecell{Norm.\\AxM Power} \\ \midrule
\multirow{4}{*}{\makecell{DeiT-S\\(8-bit AccMul\\acc. 79.34\%)}} & Ours ($\lambda$=$1$) & 76.78\% & 82.53\%         \\
                    & Ours ($\lambda$=$100$) & 76.31\% & 66.08\%         \\
                    & Evo\_1KV9         & 70.16\% & 93.17\%         \\
                    & Evo\_1L2H         & 67.01\% & 73.04\%         \\\midrule
\multirow{4}{*}{\makecell{Swin-S\\(8-bit AccMul\\acc. 81.83\%)}} & Ours ($\lambda$=$1$) & 79.54\% & 82.53\%         \\
                    & Ours ($\lambda$=$100$) & 79.15\% & 66.08\%         \\
                    & Evo\_1KV9         & 79.25\% & 93.17\%         \\
                    & Evo\_1L2H         & 76.64\% & 73.04\%         \\ \bottomrule
\end{tabular}
\end{table}

Table~\ref{tab:vit} compares \METHOD{}-generated AxMs with the Evo\_1KV9 and Evo\_1L2H AxMs from the EvoApprox library.
We copy the accuracy results of these designs from the TransAxx paper
and measure their power using the ASAP 7nm library.
Other AxMs tested in TransAxx are excluded due to their impractically large accuracy loss.
Table~\ref{tab:vit} suggests that \METHOD{} provides a favorable accuracy-power trade-off under our setting.
For DeiT-S and Swin-S, 
using $\lambda=1$ keeps the accuracy acceptable compared to 
that of the 8-bit quantization model using AccMuls,
while reducing AxM power by 17.47\%.
Increasing $\lambda$ to $100$ provides further power reduction with a small decrease in accuracy. 
In contrast, Evo\_1KV9 consumes much more power,
while Evo\_1L2H achieves lower power at the cost of a much larger accuracy loss. 
For Swin-S, comparing our AxM when $\lambda=100$ with Evo\_1KV9,
\METHOD{} achieves comparable accuracy and reduces power by an additional 27.09\%.

\begin{table}[tb]
\centering
\caption{Comparison of ResNet50 accuracy, normalized AxM energy, and runtime using different layer-wise AxM exploration methods on CIFAR-10.
``N/A'' means not applicable.}
\label{tab:mixed-type}

\begin{tabular}{lrrr}
\toprule
Method         & \makecell{ResNet50\\Accuracy} & \makecell{Norm. \\Energy} & \makecell{Runtime\\/hour} \\ \midrule\midrule
FP32               & 93.65\% & N/A                 & N/A          \\
AccMul             & 93.56\% & 100.00\%            & N/A          \\ \midrule
Ours ($\lambda$=1)    & 93.71\% & 74.97\%                  & 0.69         \\
Ours ($\lambda$=10)   & 93.65\% & 55.64\%                  & 0.70         \\
Ours ($\lambda$=100)  & 92.79\% & 35.90\%                  & 0.69         \\
Ours ($\lambda$=1000) & 92.97\% & 35.81\%                  & 0.69         \\ \midrule
MARLIN-1           & 92.14\% & 80.67\%             & 111.1        \\
MARLIN-2           & 91.70\% & 76.67\%             & 111.1        \\ \midrule
ALWANN-1           & 89.08\% & 78.47\%             & N/A          \\
ALWANN-2           & 88.58\% & 70.02\%             & N/A          \\ \bottomrule
\end{tabular}

\end{table}

\subsection{Experiments with Layer-Wise Different AxM Types}\label{subsect:exp-sota}



This set of experiments assumes that different layers can use different types of AxMs.
To support this in \METHOD{},
we allow structure parameters to differ across layers,
\ie{},
$\Theta^{(1)}$, $\Theta^{(2)}$, $\ldots$, $\Theta^{(L)}$ can be different.
We compare \METHOD{} with existing mixed-type AxM selection frameworks for CNNs,
\ie{}, MARLIN~\cite{guella2024marlin} and ALWANN~\cite{mrazek2019alwann}.
Both MARLIN and ALWANN select different AxM types across layers to improve energy efficiency.
We use data reported in the MARLIN paper,
which also includes results for ALWANN.
Although MARLIN is evaluated in a 65 nm technology and our evaluation is based on the ASAP 7nm library, 
the AxM energy normalized to the AccMul energy still provides a meaningful basis for comparison.

Table~\ref{tab:mixed-type} compares \METHOD{} with MARLIN and ALWANN on CIFAR-10 
using the ResNet50 model and the w8a8 quantization scheme.
The AxM energy consumption refers to the energy consumed by all AxMs during inference of one input image.
It is estimated by accumulating the energy of approximate multiplications in convolutional layers.
The normalized AxM energy is estimated from number of multiplications and AxM power in each layer.
Since the tested 8-bit multipliers have nearly identical latency,
we assume the same delay for all multipliers,
so the delay term cancels out in the normalization.
Therefore, the normalized energy is computed from $\sum_l \#\tit{mults}^{(l)} \times \tit{AxMPower}^{(l)}$,
normalized to the 8-bit AccMul.

From Table~\ref{tab:mixed-type},
our method consistently outperforms both MARLIN and ALWANN in terms of accuracy and energy efficiency.
When $\lambda=1$,
our method reaches 93.71\% accuracy,
which slightly exceeds the baseline 8-bit AccMul accuracy of 93.56\%.
Meanwhile,
AxM energy consumption is reduced by 25.03\%.
When $\lambda=1000$,
our method achieves 92.97\% accuracy with 35.81\% normalized energy.
Comparing the case of $\lambda=1000$ with MARLIN-2,
\METHOD{} improves accuracy by 1.27\% while reducing energy consumption by 40.86\%.
One reason for the large energy savings
is that \METHOD{} explores a larger AxM design space.
With the parameterization in Section~\ref{subsect:indicator} and the mapping in Section~\ref{sect:extract},
\METHOD{} applies constant-0 replacements to HA and FA sum and carry outputs.
In contrast, MARLIN explores a more restricted design space based on removing entire columns of partial products.
Comparing runtime,
\METHOD{} requires 0.7 hours on a single NVIDIA A100 GPU for the 20 training epochs,
while MARLIN consumes over 111 hours using 16 threads of a Ryzen 5950X CPU and an NVIDIA RTX A5000 GPU.




\section{Conclusion}\label{sect:concl}

In conclusion,
to achieve a good trade-off between power consumption and accuracy of AxM-based AI accelerators,
we propose \METHOD{} to directly explore the AxM design space through model retraining.
At the same accuracy level,
\METHOD{} significantly reduces AxM power on several CNNs and vision transformers on the CIFAR-10 and ImageNet datasets.
In the future,
we will extend \METHOD{} to large language models.

\balance
\bibliographystyle{unsrt}
\bibliography{ref.bib}

\end{document}